\documentclass[transmag,letterpaper]{IEEEtran}
\usepackage{nopageno}
\newif\ifpeerreview

\peerreviewfalse

\newcommand{\paperID}{12}

\usepackage[switch]{lineno}

\usepackage{cite}
\usepackage{url}
\usepackage{amsmath}
\usepackage{lipsum}
\usepackage{booktabs} 
\usepackage{colortbl}
\usepackage{subfigure}
\usepackage{enumitem}
\usepackage[linesnumbered,algoruled,boxed,lined]{algorithm2e}
\usepackage{nicefrac}
\usepackage{xspace}
\usepackage{float}
\usepackage{placeins}
\usepackage{bm}
\usepackage{siunitx}
\usepackage{graphicx}
\graphicspath{ {./images/} }

\definecolor{Yellow}{rgb}{1,1, 0.6}
\definecolor{Red}{rgb}{1, 0.6, 0.6}
\definecolor{Blue1}{rgb}{0, 0.6, 1}
\definecolor{green}{rgb}{0.2, 0.8, 0.2}
\definecolor{Red1}{rgb}{0.5, 0.2, 0.2}
\definecolor{PaleYellow}{rgb}{0.8,0.8, 0}
\definecolor{Purple}{rgb}{1.0, 0, 1.0}

\newcommand{\TODO}[1]{}
\newcommand{\todo}[1]{}

\newcommand{\etal}{\textit{et al}.}
\newcommand{\ie}{i.e.\xspace}
\newcommand{\eg}{e.g.\xspace}
\newcommand{\Mod}[1]{\ (\mathrm{mod}\ #1)}

\ifCLASSINFOpdf
\else
\fi

\begin{document}

\ifpeerreview
  \linenumbers
  \linenumbersep 5pt\relax
\fi 

%
\title{Wireless Software Synchronization of Multiple Distributed Cameras}


\ifpeerreview
\author{Anonymous ICCP 2019 submission \\
Paper ID \paperID}
\else
\author{\IEEEauthorblockN{Sameer Ansari,
Neal Wadhwa, 
Rahul Garg, 
and Jiawen Chen}
\IEEEauthorblockA{Google Research, Mountain View, CA}
}
\fi

\ifpeerreview
\markboth{Anonymous ICCP 2019 submission ID \paperID}%
{}
\else
\fi

\IEEEtitleabstractindextext{%
\begin{abstract}
We present a method for precisely time-synchronizing the capture of image sequences from a collection of smartphone cameras connected over WiFi.
Our method is entirely software-based, has only modest hardware requirements, and achieves an accuracy of less than $\SI{250}{\micro\second}$ on unmodified commodity hardware. It does not use image content and synchronizes cameras prior to capture.
The algorithm operates in two stages.
In the first stage, we designate one device as the leader and synchronize each client device's clock to it by estimating network delay.
Once clocks are synchronized, the second stage initiates continuous image streaming, estimates the relative phase of image timestamps between each client and the leader, and shifts the streams into alignment.
We quantitatively validate our results on a multi-camera rig imaging a high-precision LED array and qualitatively demonstrate significant improvements to multi-view stereo depth estimation and stitching of dynamic scenes.
We release as open source \texttt{libsoftwaresync}, an Android implementation of our system, to inspire new types of collective capture applications.
\end{abstract}

\ifpeerreview
\else
\fi
}

\maketitle
\thispagestyle{empty}

\IEEEdisplaynontitleabstractindextext

\section{Introduction}

\begin{figure}[!t]
    \centering
    \includegraphics[width=\linewidth]{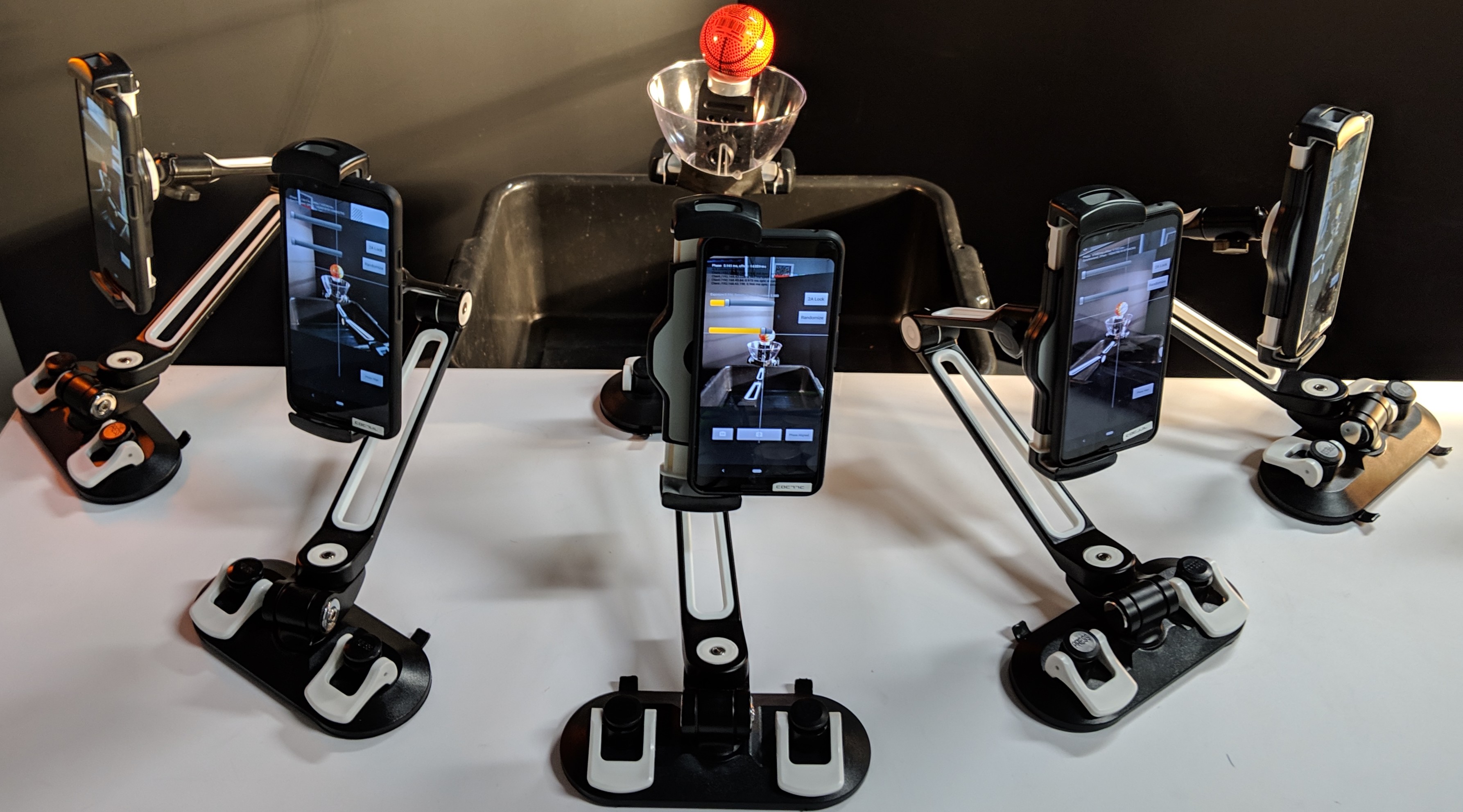}
    \caption{Five smartphone cameras simultaneously capture a highly dynamic scene. The cameras are synchronized using our software-based solution, which requires no wires or additional hardware. The images are shown in Fig. \ref{fig:teaser}.}
    \label{fig:teaser_setup}
\end{figure}

Many computer vision algorithms require multiple images from different viewpoints as input. These algorithms can fail spectacularly when applied to dynamic scenes if the images are not captured at the same time. Depth from stereo, view interpolation, and view supervision losses are just a few examples of such algorithms. We provide a solution to temporally synchronize the cameras of multiple inexpensive smartphones, so that they can capture images simultaneously (Fig.~\ref{fig:teaser_setup}).  

Our solution is software-based, does not require any additional hardware, and scales up to multiple devices. Furthermore, the devices are synchronized prior to capture and can be located in any physical arrangement as long as they can communicate over a wireless network.


Traditionally, synchronizing cameras in advance of capture is solved by using specialized hardware such as analog video gen-lock or the IEEE 1394 isochronous interface. While effective and reliable, hardware solutions require cumbersome wires, limiting their portability, and more importantly, are unavailable on the vast majority of cameras. For non-specialized cameras, this limits usage to nearly static scenes where several frames of timing error is acceptable.

Another option is to capture image sequences and align the frames after capture. A clapperboard or similar ``oracle'' device that is visible to all cameras can assist in aligning the frames. However, accuracy is limited to half a frame duration. Post-processing methods to obtain sub-frame alignment exist~\cite{bradley2009synchronization, vsmid2017rolling,padua2010linear,tresadern2003synchronizing,dai2006subframe}, but can break down for low-texture or noisy scenes where finding correspondences is hard. 
Even if these algorithms yield perfect sub-frame alignment, it is necessary to temporally interpolate the frames to the aligned sub-frame time~\cite{padua2010linear}, which is another challenging problem. In addition, such methods may require capturing much more data than necessary to ensure that the same instant in time is captured by all cameras.

\begin{figure*}
    \centering
    \includegraphics[width=\linewidth]{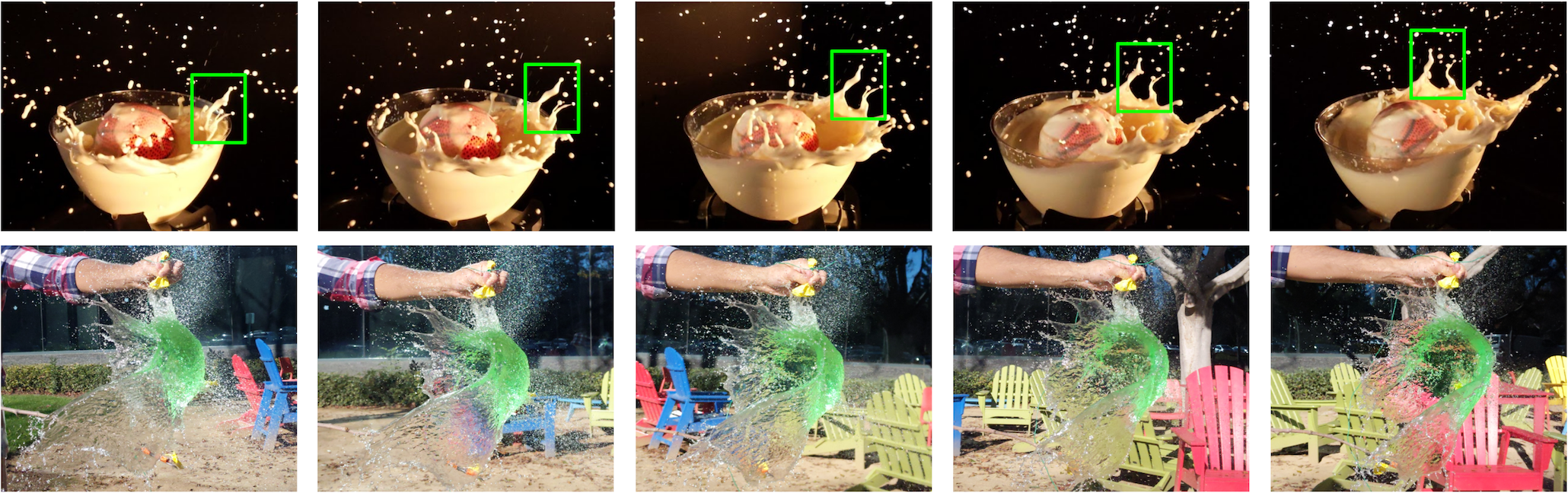}
    \caption{Top: Synchronized images of a ball splashing in milk captured by the cameras in Fig.~\ref{fig:teaser_setup}. The same splash of milk is highlighted in green in the photos. Bottom: Synchronized outdoor images of a bursting water balloon with a wider baseline setup. Note the consistent shape of the water concavity.}
    \label{fig:teaser}
\end{figure*}

In contrast, our system is capable of achieving a synchronization accuracy of less than $\SI{250}{\micro\second}$ before capture, allowing for the multi-view simultaneous capture of highly dynamic phenomena such as sports action shots, birds in flight, and splashing liquids (Fig.~\ref{fig:teaser}). To acquire synchronized images, users are only required to install our software, position the phones, and press the shutter button on one of the phones. Our system does not look at image content; instead, it uses only image timestamps and wireless messages for synchronization. This means the cameras can have non-overlapping fields of view and can be placed in any configuration.  For example, a hand-held, inexpensive light field capture rig can be created by placing the cameras in a square array. Or the cameras can be placed in a linear array, so that a view interpolation algorithm can create a ``bullet time'' effect where the camera moves around an apparently frozen dynamic subject~\cite{WangWC15,Wang:2016:PCS:2856767.2856778}. 

Our system is especially useful for collecting data for deep learning algorithms that require synchronized images for training \cite{Xie16,Garg16, Godard17} and also need to be deployed on a smartphone. Were such data collected with hardware-synchronized non-smartphone cameras, biases introduced by differences in the sensor or lens could ruin the algorithm's performance. A portable rig containing smartphones synchronized using our system provides an inexpensive method to collect such datasets in the wild.

Our system operates in two stages. First, the clocks of all the phones are synchronized to the clock of a leader device using a variant of the Network Time Protocol (NTP). This synchronization is only enough to achieve an accuracy of half a frame duration (\eg, $\SI{16}{\milli\second}$). In the second stage, the cameras are instructed to capture a continuous stream of images. We then shift the phase of all client streams to that of the leader and achieve an accuracy better than $\SI{250}{\micro\second}$.

In summary, we contribute the following:
\begin{itemize}
    \item Two algorithms for efficiently phase aligning camera image streams to arbitrary precision.
    \item A system combining phase alignment with a variant of NTP that achieves an accuracy better than $\SI{250}{\micro\second}$ on unmodified consumer smartphones over WiFi.
    \item A theoretical analysis and empirical confirmation of the accuracy and limits of our system.
\end{itemize}
We release as open source \texttt{libsoftwaresync}, an Android implementation of our system.

\section{Related Work}
\label{sec:related_work}
Precision time synchronization in a distributed system is a classically hard problem.
The fact that each device is physically independent with its own hardware clock makes synchronization necessary even if networking were reliable with low communications latency. When applied to the camera synchronization problem, misalignment is primarily due to two factors. First, each device has an independent notion of the current time, which may drift and is at best only loosely aligned to an external source (\eg, GPS or cell network). Second, consumer cameras typically do not feature a mechanism by which image capture can be triggered to occur at a specific time. Instead, there is a large firmware and software stack between the camera hardware and the application that requests an image, which introduces additional variable latency.
Without physical wires to synchronize all clocks and cameras in the network, it is challenging to trigger all cameras at the same time.

A number of systems focus on aligning video sequences and correcting for rolling shutter after they have been captured. Some systems use active illumination to tag frames from which an offset may be inferred~\cite{bradley2009synchronization}. Others rely on the scene content itself, by detecting abrupt lighting changes~\cite{vsmid2017rolling}, tracking continuous periodic motion~\cite{padua2010linear}, or requiring motion be simple enough that the spatiotemporal transformation can be represented by a low-dimensional model~\cite{tresadern2003synchronizing, dai2006subframe}. As the source imagery is fundamentally unsynchronized, these post-capture techniques all ultimately rely on optical flow and interpolation to bring images into sub-frame alignment. Our goal is to synchronize cameras \emph{before} capture, ensuring they expose the same instant in time and sidestepping the frame interpolation problem.


Gu \etal~\cite{HerakleousP14} describe a simple scheme for projector-camera synchronization by forcing the camera to wait until the projector has displayed an image before triggering capture.
This technique is straightforward to implement but cannot capture moving subjects, can introduce unacceptably long shutter lag, and is impractical for video sequences.
Litos \etal~\cite{litos2006synchronous} describe a system that uses the Network Time Protocol (NTP) \cite{Mills:2010:CNT:1951866} to align device clocks but neglects the variable latency between when software sends a shutter command and when the camera ultimately captures a frame.
Petkov\'{c} \etal~\cite{petkovic2016software} describe a clever system for synchronizing a projector and a camera by using the camera itself to measure the signal propagation delay. But this requires a projector and also does not account for the variable latency in triggering a camera. 

The system by Ahrenberg \etal~\cite{ahrenberg2004mobile} is closest in spirit to ours. Like our system, they rely on NTP to synchronize device clocks before sending a single recording command. Their system relies on the IEEE 1394 hardware trigger to keep pace, but like the other systems above, assumes that the delay between the software trigger command and image capture to be the same among all cameras. Our system does not require a hardware trigger and explicitly models the variable latency between the software trigger command and image exposure.

\section{Overview}
\label{sec:overview}


We give a high level overview of our system.
We first list our assumptions of the underlying system, which justify our algorithm and where it may be deployed.

\subsection{Setting}
Our system consists of a collection of $N$ \emph{devices}, each of which is equipped with a camera and a network card.
In practice, these are smartphones with an integrated camera and WiFi.
At startup, the devices connect over WiFi, synchronize their clocks, and puts their cameras into continuous streaming mode (Sec.~\ref{sec:sync_device_clocks}).
After a second stage where camera streams are phase aligned (Sec.~\ref{sec:phase_align}), initialization is complete.
At any time after, when a capture request containing a timestamp arrives, all devices simply save the appropriate images from a ring buffer to disk. Note that this requested timestamp can be in the past, enabling zero shutter lag captures.
Our goal is to do the above with reasonable latency and as such, rely on a few assumptions of the underlying hardware:
\begin{enumerate}
    \item \textbf{Modest variance in network latency}.
    The first stage of our algorithm brings the devices' \emph{local clocks} into alignment.
    Theoretically, our system can tolerate arbitrary latency variance, although it may take an unacceptable amount of time to converge.
    To support arbitrary devices without hardware timestamping of network packets, we include in our latency model variable delays induced by the operating system's networking stack.
    \item \textbf{Hardware camera timestamps}.
    We require that the camera hardware tag images with timestamps \emph{in the domain} of the device's local clock.
    A key component of our algorithm is accurately modeling the time between requesting an image and when it is captured.
    If images were not timestamped in the same clock domain, this modeling would not be possible.
    iPhones and numerous Android devices support this feature. 
    \item \textbf{Hardware image streaming}
    We require that the camera hardware or firmware be able to latch capture settings such that it can stream frames indefinitely with low timing variance between frames.
    For example, if we request images at ISO 100, exposure time of $\SI{10}{\milli\second}$, and a frame duration (time between frames) of $\SI{33}{\milli\second}$, we should expect a stream of images with timestamps close to $\SI{33}{\milli\second}$ apart.
    Most commercially available cameras and smartphones support this feature.
\end{enumerate}

\section{Synchronizing Device Clocks}
\label{sec:sync_device_clocks}

Like previous work\cite{litos2006synchronous,ahrenberg2004mobile}, we use a variant of the Network Time Protocol (NTP) \cite{Mills:2010:CNT:1951866} to synchronize device clocks, for which we give a simplified overview. This step is unnecessary if all devices had support for the Precision Time Protocol (PTP) \cite{ptp_ieee_std}, which provides sub-microsecond hardware synchronization. However, PTP hardware is typically not available on consumer smartphones. In our system, the user designates one device as the leader and creates a WiFi hotspot. The remaining $N-1$ clients connect to the leader and estimate their clock offsets. Typically, device clock synchronization is only performed once at the beginning of a capture session. However, depending on the application (\eg, very long sessions such as time lapses), one may want to periodically re-synchronize clocks to account for clock drift or exploit better WiFi conditions. In our experiments, clocks remained stable over the course of one hour.

\begin{figure}
    \centering
    \includegraphics[width=\linewidth]{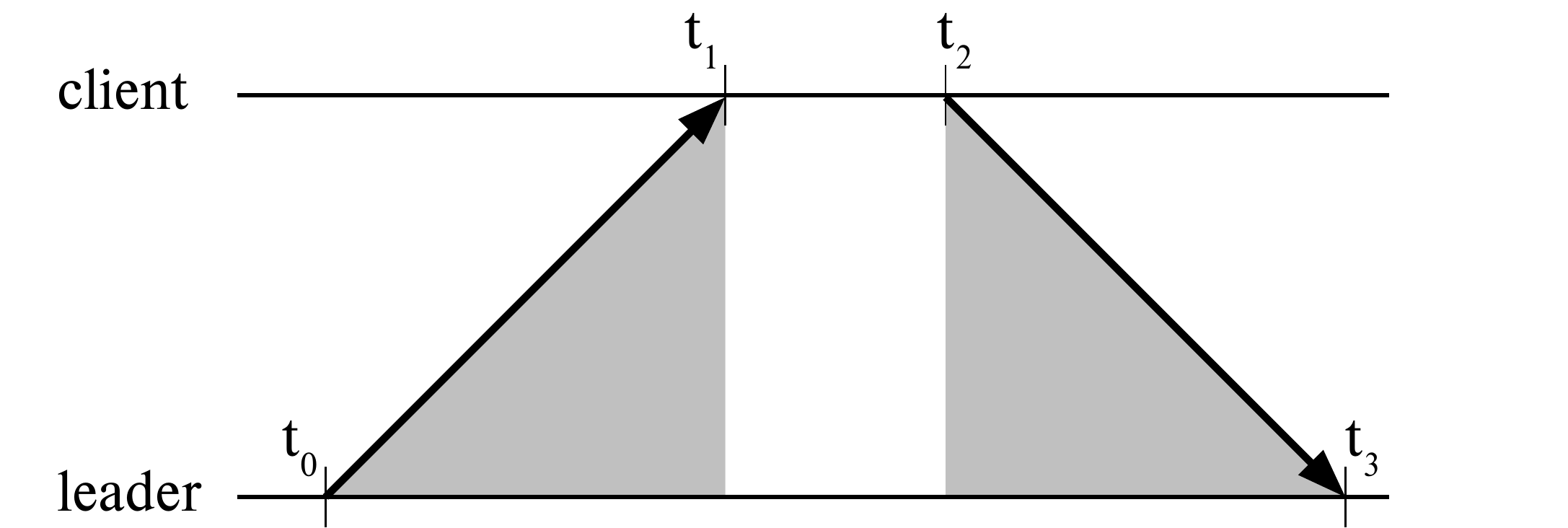}
    \caption{NTP handshake. With one message in each direction, the leader can estimate the clock offset $\theta$.}
    \label{fig:ntp}
\end{figure}

A single message in the synchronization routine consists of the following handshake (Fig.~\ref{fig:ntp}) between the leader and one of the client devices (multiple clients are handled independently):

\begin{enumerate}
  \item At time $t_0$ in the leader's clock domain the leader sends the message ``$t_0$''.
  \item At time $t_1$ in the client's clock domain, it receives the message ``$t_0$''.
  \item At time $t_2 > t_1$ in the client's clock domain, it sends the message ``$(t_0, t_1, t_2)$''.
  \item At time $t_3 > t_0$ in the leader's clock domain, it receives the message ``$(t_0, t_1, t_2)$''.
\end{enumerate}

We then estimate clock offset $\theta$ and round-trip delay $\phi$:


\begin{align}
\theta &= \frac{(t_1 - t_0) + (t_2 - t_3)}{2} & 
\phi &= t_3 - t_2 + t_1 - t_0
\end{align}



Note that the computation of $\theta$ assumes that communications latency is symmetric, as is typical in networking. If not, there will be a systematic bias which limits our overall synchronization accuracy. 
Since individual clock readings and network latency measurements may be noisy (due to \eg, buffering), we collect multiple samples and filter them. To synchronize $N$ devices, each of the $N-1$ client devices exchanges $K$ messages with the leader in a round robin fashion. Following the best practices suggested by NTP, we determine $K$ using the mean and min filters.

\subsection{Mean Filter}
The mean filter assumes that communications latency is normally distributed. This lets us compute the sample mean and variance after rejecting outliers where the round-trip delay is greater than some threshold (\eg, due to buffering or interference). Since variance decreases as $1/K$, we can determine the number of messages $K$ required to reach a desired accuracy.
However, in our experiments, we found that the mean filter converges to a systematic bias of $\approx \SI{0.4}{\milli\second}$ when compared to an external reference clock. To ameliorate this bias, we use the min filter.

\subsection{Min Filter}
The min filter assumes that the sample with the \emph{minimum} latency is the most reliable. This heuristic, also used by the NTP standard, is based on the hypothesis that the message with the shortest delay experienced the least amount of processing and therefore is the most symmetric. We confirm this hypothesis quantitatively and compare the mean and min filters in Sec.~\ref{sec:results}.
In our implementation, we can either exchange a fixed number of messages and take the minimum, or iterate until we observe a sample with latency below some threshold. To guarantee a predictable startup time of less than 10 seconds on our rig of 5 smartphones with a mean latency of $\SI{4}{\milli\second}$, we used a fixed $K=300$ samples. 

\section{Phase-based image stream alignment}
\label{sec:phase_align}

Once all device clocks are synchronized, we can assume that all images are timestamped in the leader's clock domain (by adding the appropriate offset). However, to actually capture images simultaneously on all devices, we need to account for the latency between when software on a device requests a frame and when the sensor is actually exposed. This latency can be highly variable and is hard to measure accurately. Instead, we exploit hardware image streaming: the ability to indefinitely capture a sequence of images at regular intervals. We seek to shift the phases of each stream, so that all cameras are exposing at the same time. Then, all that remains is to select from the image ring buffer which frames to save to disk.

Suppose that the leader camera and a client camera start streaming at times $u_{goal}$ and $u$ respectively and repeatedly capture images with period $T$.
We seek to reduce the difference in their phases  $\delta=u-u_{goal} \Mod T$ to a small value $\epsilon$ by shifting the phase of the client camera (Fig.~\ref{fig:phase_alignment}a). 
As opposed to trying to align $u$ and $u_{goal}$ directly, phase shifting a frame stream can be done efficiently and accurately.

We describe two methods to phase shift a frame stream, \emph{reset sampling}, a slower to converge method that is applicable to many devices, and \emph{frame injection}, a faster to converge method that requires device-specific modeling.



\subsection{Reset Sampling}

\begin{figure}
    \centering
    \includegraphics[width=\linewidth]{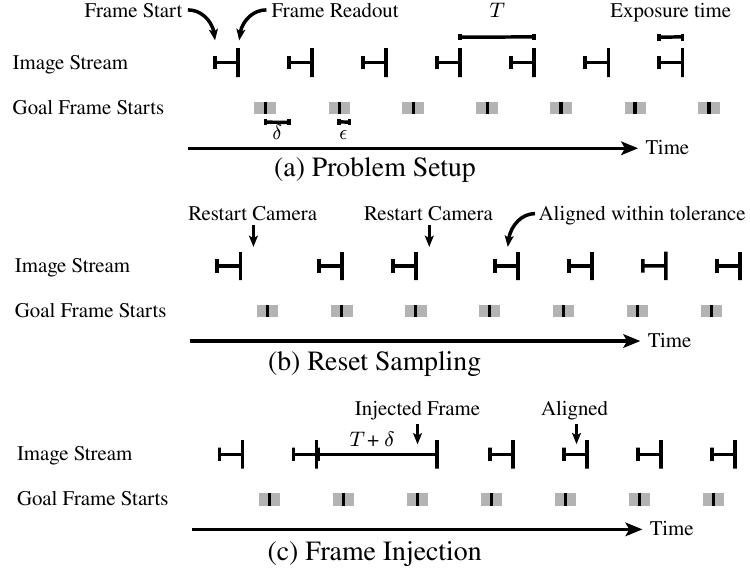}
    \caption{Different strategies to phase align a hardware image stream to a specified goal. (a) Frames are read out every $T$ ms and the start of each frame is offset from the goal by a phase of $\delta$ ms. We seek to align frame start times to within $\epsilon$ ms of the goal phase. (b) Reset sampling restarts the camera until the frame start times falls within the gray tolerance. (c) For certain cameras, we can achieve much quicker alignment by injecting a frame with exposure $T+\delta$ to directly phase shift the image stream.}
    \label{fig:phase_alignment}
\end{figure}

In reset sampling, we restart the camera until its phase falls within a tolerance $\epsilon$ of the goal phase (the gray region in Fig.~\ref{fig:phase_alignment}b). There is significant variance in how long it takes a camera to reset ($\SI{600}-\SI{800}{\milli\second}$ in our experiments). 
Furthermore, this distribution can vary from camera to camera and in the worst case can result in phases always falling outside the tolerance. Therefore, we sleep for a random amount of time $U (0, L)$ after stopping the camera and before restarting it to make the overall latency distribution more uniform. If the latency of starting the camera is described by the random variable $S$, then, the total phase offset can be modeled as:

\begin{equation}
U(0, L) + S \Mod T \approx U(0, L) \Mod{T} \approx U(0, T),
\end{equation}

where the approximation holds as long as $L$ is large relative to $T$ and the range of $S$. We found a value of $L=1$ second to be reasonable.
Since we are able to shift the phase by a random amount in each iteration, we effectively sample phase offsets from a uniform distribution and reject those that are outside the gray region in Fig.~\ref{fig:phase_alignment}b. There is an $\epsilon/T$ chance a sample will be accepted on any one iteration. This means to have a 95\% chance of converging within a tolerance of $\epsilon$, we need to reset the camera $R$ times:

\begin{equation}
\label{eq:reset_95}
R = \frac{\log{0.05}}{\log{(1-\epsilon/T)}}.
\end{equation}

For $\epsilon=\SI{1}{\milli\second}$ and $T=\SI{33}{\milli\second}$, $R=98$ iterations are required.

\subsection{Frame Injection}
Some camera APIs (\eg, Android's Camera2 API \cite{camera2}) allow a high priority capture request to be injected into a normal priority continuous image stream. Typically, the continuous stream is used for the viewfinder, while the high priority request is used to capture a still image with different exposure or gain settings. By injecting a frame with exposure longer than the duration $T$, we can shift the phase of a stream in \emph{one} iteration. In an ideal camera, if the phase offset between two image streams is $\delta$, we can align the streams by injecting a frame with exposure time $T+\delta$. Real cameras can behave differently than this ideal, but as long as the camera is deterministic, we can fit a function mapping a desired phase offset $\delta$ to an exposure that will shift the camera's stream by $kT+\delta$ for some integer $k$. For example, on one specific phone model, we empirically found that exposures of length $T + \delta/2$ would shift the phase by $2T+\delta$. Some cameras require exposure times to be an integer multiple of a scanline's readout time. While frame injection is deterministic, we model this quantization as Gaussian noise with standard deviation $\sigma$ and measured $\sigma$ to be on the order of tens of microseconds. Under this assumption, this method will converge in $4 \sigma^2/\epsilon^2$ iterations with 95\% probability. If $\epsilon=\SI{1}{\milli\second}$, frame injection succeeds in one iteration.

\subsection{Error Detection and Recovery}
Since all devices are in the same clock domain, we can detect phase misalignment directly from frame timestamps. This lets us detect frame drops or deviations in the stream period and notify the application to realign automatically. One class of errors we did not encounter in practice but would like to handle gracefully in the future is low-frequency clock drift. We can detect clock drift by periodically exchanging NTP messages, and then rerun phase alignment on the subset of devices that have drifted.


\section{Results}
\label{sec:results}
We quantitatively verify the accuracy of our method and show that it outperforms several common naive approaches by three orders of magnitude. We also demonstrate how our method qualitatively improves the quality of multi-view stereo and panorama stitching. Finally, we show simultaneous multi-view captures of several highly dynamic scenes.

\subsection{Measuring Overall Synchronization Error}

\begin{figure}
  \includegraphics[width=\linewidth]{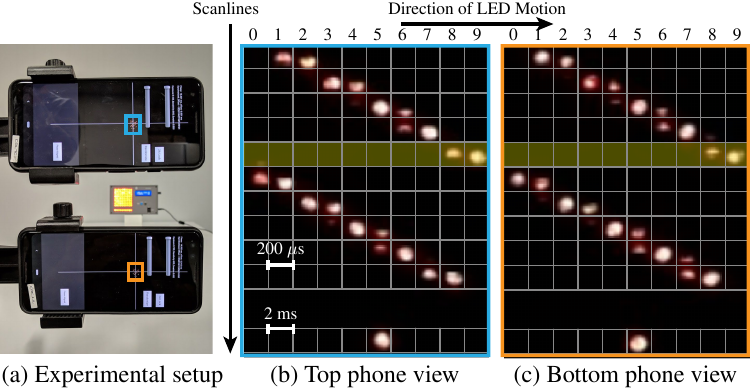}
    \caption{Ground truth measurement of synchronization accuracy. (a) Two smartphones capture images of a $10 \times 10$ LED panel with an exposure time of $\SI{100}{\micro\second}$. LEDs sweep horizontally across the grid at a rate of one column per $\SI{200}{\micro\second}$. The separate bottom row of LEDs sweep at a rate of one column per $\SI{2}{\milli\second}$.
    The central scanlines of both phones are set to capture the fifth row (highlighted in yellow).
    The skewed pattern is due to rolling shutter, where it takes $\SI{3.3}{\milli\second}$ to capture the fraction of the scene shown. (b) Top phone view of the LED panel. (c) Bottom phone view of the LED panel.}
    \label{fig:led_panel}
\end{figure}

To quantitatively validate our method, we use an LED panel as an external high-precision frequency reference (Fig.~\ref{fig:led_panel}a). The panel contains a fast-changing 10$\times$10 array and a slow-changing 10$\times$1 array. In the fast-changing array, a column is lit for a fixed time period $\tau$, then the next column is lit for $\tau$, and so forth until the last column is reached and the pattern repeats. In the slow-changing array, each LED stays on for $10\tau$. The two arrays allow us to measure synchronization error with resolution on the order of $\tau$. Since the entire panel repeats with a period of $100\tau$, it is possible though unlikely for the measured synchronization to be off by exactly $100k\tau$ where $k$ is an integer. We can eliminate this possibility by using multiple values of $\tau$.

In our first experiment, we placed two Google Pixel 3 smartphones on a tripod and captured $\SI{100}{\micro\second}$ exposures of the LED panel with $\tau$ set to $\SI{200}{\micro\second}$ (Fig.~ \ref{fig:led_panel}a). The frame duration $T$ was $\SI{33}{\milli\second}$. To avoid confounding synchronization error with misalignment caused by mismatched rolling shutters, we oriented the phones so that their center scanlines both capture the fifth row of the fast moving array of the LED panel.

We measured the synchronization error of our min filter and frame injection methods by conducting 239 trials. In each trial, we first reset the synchronization process to undo clock sync and added a random length long exposure frame to undo the phase alignment. We then asked our system to synchronize to a tolerance of $\epsilon = \SI{20}{\micro\second}$. Example captures from the two phones are shown in Fig.~\ref{fig:led_panel}(b-c). The skewed pattern of LEDs is due to rolling shutter.

In all pairs of images, there is no more than 1 LED difference between the pairs, showing that the total synchronization error is less than $\SI{200}{\micro\second}$. We also ran experiments with $\tau$ set to $\SI{10}{\milli\second}$ to verify that our results were not off by an integer multiple of $\SI{20}{\milli\second}$.
To get more fine-grained error measurements, we exploit the fact that short exposures and rolling shutter means some LEDs will only be partially captured. This allows us to estimate the error between the two phones with temporal resolution better than $\tau$. We use the mean pixel intensity in each grid cell in the fifth row of Fig.~\ref{fig:led_panel}(b-c) to estimate the sub-grid position of the LED lights.

For two devices, we found the maximum error between them to be $\SI{121}{\micro\second}$ (Fig.~\ref{fig:TimingHistograms}c and Table.~\ref{table:naive_vs_ours}). With more than two devices, any pair can have this maximum error but in opposite directions implying a worst case error of $\SI{242}{\micro\second}$.

We also compared to three naive synchronization methods: wired, Bluetooth, and WiFi. In the \emph{naive wired} method, we used a selfie stick connected via an audio splitter to two Pixel 3 smartphones, such that a single button press would trigger both phones. We used the same setup as in Fig.~\ref{fig:led_panel}(a), except we set $\tau$ to $\SI{20}{\milli\second}$. In our \emph{naive Bluetooth} method, we connected wireless Bluetooth photo capture triggers to each device and used a mechanical button to press both triggers at the same time. In our \emph{naive WiFi} method, we connected one phone to the other's WiFi hotspot. On shutter press, the phone with the hotspot then instructed itself and the client to take photos immediately. We conducted 20 trials for each of these methods. In the wired and Bluetooth methods, 2 and 1 shot failed to capture respectively due to issues with the extra hardware. These naive methods have errors over 1000 times our method and occasionally fail to work (Table~\ref{table:naive_vs_ours}). In addition, the wired and Bluetooth solutions require extra hardware making them cumbersome to use and difficult to scale to more devices.

\begin{table}
    \centering
    \caption{Synchronization Accuracy of Naive vs Our Method}
    \begin{tabular}{rrrr}
    \hline
    \multicolumn{1}{c}{\textbf{Method}} & \multicolumn{1}{c}{\textbf{Max error}} & \multicolumn{1}{c}{\textbf{Mean Abs Error}} & \multicolumn{1}{c}{\textbf{Stdev}} \\ \hline
    Naive Wired & $\SI{180}{\milli\second}$ & $\SI{103}{\milli\second}$ & $\SI{50}{\milli\second}$ \\ 
    Naive Bluetooth & $\SI{200}{\milli\second}$ & $\SI{69}{\milli\second}$ & $\SI{65}{\milli\second}$ \\ 
    Naive WiFi & $\SI{677}{\milli\second}$ & $\SI{123}{\milli\second}$ & $\SI{84}{\milli\second}$ \\ \hline 
    \textbf{Our Method} & \textbf{$\SI{0.121}{\milli\second}$} & \textbf{$\SI{0.032}{\milli\second}$} & \textbf{$\SI{0.025}{\milli\second}$} \\ \hline 
    \end{tabular}
    \label{table:naive_vs_ours}
\end{table}



\subsection{Measuring Clock and Phase Accuracy} 
The overall accuracy of our system, $\epsilon_{total}$, depends on many factors. The two main components are due to errors in clock estimation ($\epsilon_{clock}$) and errors in image stream phase alignment ($\epsilon_{phase}$). There are also other sources which we do not account for, such as imperfect hardware timestamps and variances in the readout period $T$. Ignoring error from other sources, we model the total error as:

\begin{equation}
\epsilon_{total} = \epsilon_{clock} + \epsilon_{phase}.
\end{equation}

In the previous section, we directly measured $\epsilon_{total}$ using an external reference. We can also directly measure $\epsilon_{phase}$ by comparing an image's hardware timestamps to what was requested. $\epsilon_{clock}$ can then be estimated as $\epsilon_{total} - \epsilon_{phase}$.


\begin{figure}
    \centering
    \subfigure[Phase Error]{\includegraphics[width=0.32\linewidth]{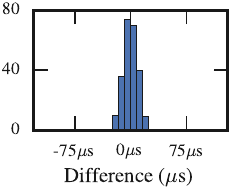}}
    \subfigure[Clock Error]{\includegraphics[width=0.32\linewidth]{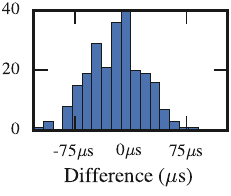}}
    \subfigure[Total Error]{\includegraphics[width=0.32\linewidth]{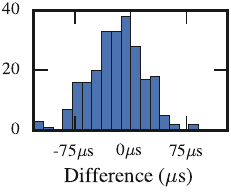}}
    \caption{Histograms of the phase alignment, network clock synchronization, and total synchronization errors over 239 trials.}
    \label{fig:TimingHistograms}
\end{figure}

We first measure phase error over the same 239 trials as in the previous section. We found that the phase error is always within our requested tolerance of $\SI{20}{\micro\second}$ and that the distribution of differences has zero mean (Fig.~\ref{fig:TimingHistograms}a). The clock error is not zero mean and has a slight negative bias (Fig.~\ref{fig:TimingHistograms}b). We analyze this further in the next section. 

Note that both the maximum and mean absolute error are comfortably less than $\SI{1}{\milli\second}$. Table~\ref{table:out_method_phase_network_total} also shows that the dominant component of the error in our system is due to clock error, which, as noted earlier, can be reduced to less than 1 microsecond by using PTP hardware.

\begin{table}[htb]
    \centering
    \caption{Distribution of Errors by Component}
    \begin{tabular}{rrrrl}
    \hline
    \multicolumn{1}{l}{\textbf{}} & \multicolumn{1}{l}{\textbf{Max error}} & \multicolumn{1}{l}{\textbf{Mean Abs Error}} & \multicolumn{1}{l}{\textbf{Stdev}} &  \\ \hline
    Clock & $\leq \SI{121}{\micro\second}$ & $\leq \SI{32}{\micro\second}$ & $\leq \SI{25}{\micro\second}$ \\ 
    Phase & $\SI{21}{\micro\second}$ & $\SI{7}{\micro\second}$ & $\SI{5}{\micro\second}$ \\ \hline
    \textbf{Total} & \textbf{\SI{121}{\micro\second}} & \textbf{\SI{32}{\micro\second}} & \textbf{\SI{25}{\micro\second}} \\
    \hline
    \end{tabular}
    \label{table:out_method_phase_network_total}
\end{table}

\subsection{Analysis of Clock Filters}



\begin{figure}
    \centering
    \includegraphics[width=3.2in]{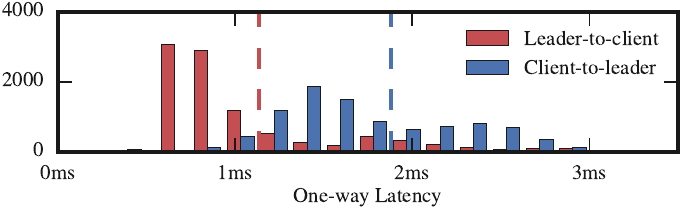}
    \caption{Histogram of one-way network latency for 10,000 messages between leader and client. The dashed lines show the means of the distributions.}
    \label{fig:systematic_bias_histogram}
\end{figure}

We analyze the behavior and effectiveness of using the mean and min clock filters.
With the same setup, the two phones exchange 10,000 NTP messages, and capture one image pair of our LED panel. From this image pair, we can estimate the true clock offset, from which we can derive the one-way latency of each message.
Like our other experiments, images are hardware timestamped and messages are software timestamped, both in each device's local clock domain. Each message takes roughly $\SI{3}{\milli\second}$ round-trip and the entire procedure takes 34.3 seconds.
Fig. \ref{fig:systematic_bias_histogram} shows the difference in one-way latency distributions over 10,000 messages after removing 64 outliers where the latency exceeds $\SI{10}{\milli\second}$. 

\begin{table}
    \centering
    \caption{Network latency symmetry and bias: Mean vs. Min}
    \begin{tabular}{rrrr}
    \hline
    \multicolumn{1}{c}{\textbf{}} & \multicolumn{1}{c}{\textbf{Mean}} & \multicolumn{1}{c}{\textbf{Min}} \\ \hline
    \textbf{Leader to client} & $\SI{1133}{\micro\second}$& $\SI{517}{\micro\second}$\\
    \textbf{Client to leader} & $\SI{1878}{\micro\second}$& $\SI{479}{\micro\second}$\\ \hline
    \textbf{Abs. latency difference} & $\SI{746}{\micro\second}$& $\SI{38}{\micro\second}$\\
    \textbf{Round-trip latency} & $\SI{3012}{\micro\second}$& $\SI{996}{\micro\second}$\\
    \textbf{NTP bias} & $\SI{372}{\micro\second}$& $\SI{19}{\micro\second}$\\ \hline
    \end{tabular}\par
    \label{table:comparing_mean_min}
\end{table}

\textbf{Mean Filter} -
NTP's clock offset accuracy relies on network symmetry. Fig.~\ref{fig:systematic_bias_histogram} and Table~\ref{table:comparing_mean_min} shows that this is not the case in our setup and that computing the mean offset over many messages converges to a bias that is approximately half the absolute latency difference.
The exact values will vary based on network configuration and hardware used. The supplement contains additional details on how we model the relationship between round-trip latency and bias.

\textbf{Min Filter} -
The min filter approach uses the message with the smallest round-trip latency. In Table~\ref{table:comparing_mean_min}, NTP bias is approximately half the difference in one-way latencies. The min filter, by choosing the sample with lowest latency, is subject to significantly less latency asymmetry and therefore achieves a much lower bias in its clock offset.
We conducted 3 quantitative experiments with sample sizes $K=300, 1000, 10000$ and found that on average, the shortest round-trip latency to be $\SI{1.87}{\milli\second}$, $\SI{1.02}{\milli\second}$, and $\SI{0.996}{\milli\second}$, respectively with low variance.
In practice, we use 300 samples which takes under 1 second per client and still provides $<\SI{1}{\milli\second}$ total synchronization accuracy. 

Errors under $\SI{200}{\micro\second}$ are likely to be confounded by measurement noise such as misaligned scanlines between phones, rolling shutter skew, and our method of using an LED panel running at $\SI{200}{\micro\second}$. Phase error is under $\SI{21}{\micro\second}$, which is minor compared to compared to the total error (Table~\ref{table:out_method_phase_network_total}).

\subsection{Phase Alignment Convergence Time}



We measure the number of iterations and total wall time required for phase alignment using both the frame injection and reset sampling methods. Note that once a client has synchronized its clock with the leader, phase alignment can proceed independent of other clients. Phase aligning $N$ devices takes only as long as the slowest device.

For the frame injection method, we used the same dataset of 239 trials as in earlier experiments. Aligning to $\epsilon = \SI{20}{\micro\second}$ took between 4-7 injected frames, with a mean of 5.5 frames and $\sigma=0.97$ frames. On Pixel 3, each injected frame takes $\SI{300}{\milli\second}$ so phase alignment converges in less than 3 seconds.

To measure reset sampling, we ran 10 additional trials with a tolerance of $\epsilon = \SI{1}{\milli\second}$. As expected, reset sampling took significantly longer than frame injection, taking on average 28.7 iterations before convergence with a standard deviation of 25.2 iterations. Because reset sampling requires restarting the camera and sleeping for a random amount ([0, 1] seconds), each iteration took on average 1.23 seconds ($\sigma = 0.27$ seconds). According to Eq.~\ref{eq:reset_95}, it will take 98 iterations or 120 seconds to achieve a 95\% probability of alignment. Although reset sampling is significantly more expensive than frame injection, it does not require any modeling of camera request latency and is deployable on a wider variety of devices.

\subsection{Applications}
\begin{figure*}
    \centering
    \includegraphics[width=0.98\linewidth]{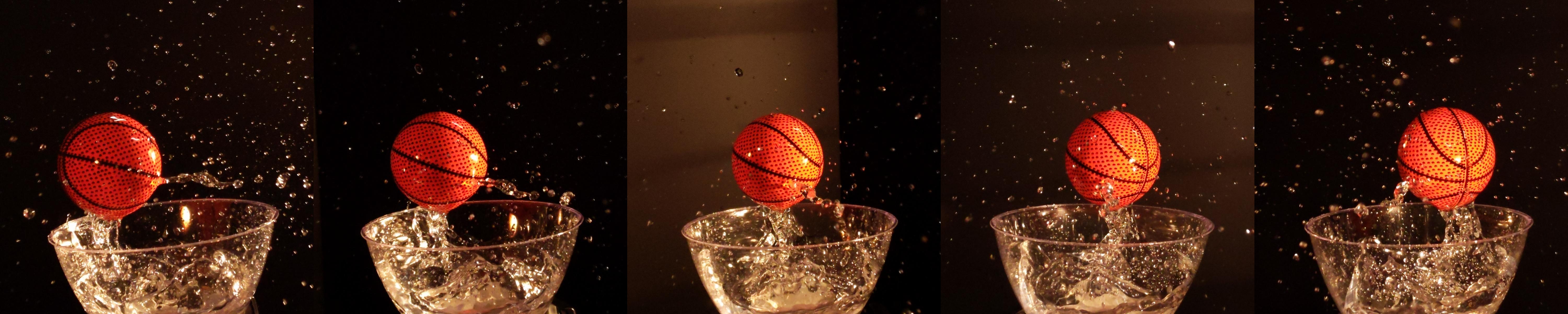} \\ \vspace{0.2mm}
    \includegraphics[width=0.98\linewidth]{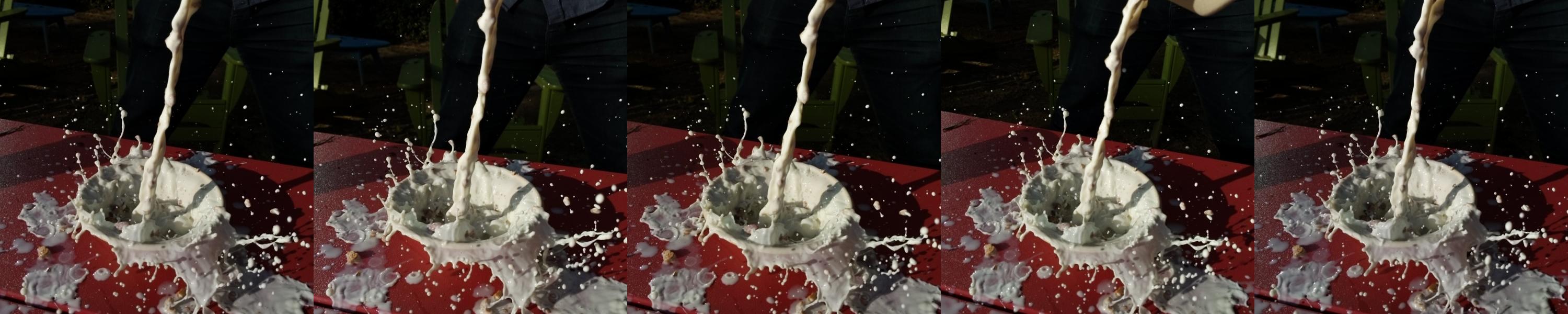} \\ \vspace{0.2mm}
    \includegraphics[width=0.98\linewidth]{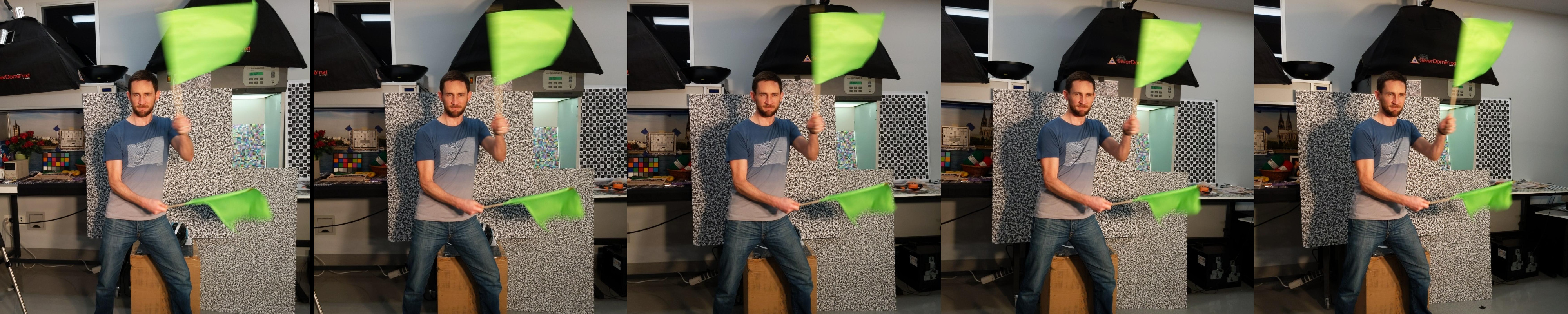} \\ \vspace{0.2mm}
    \includegraphics[width=0.98\linewidth]{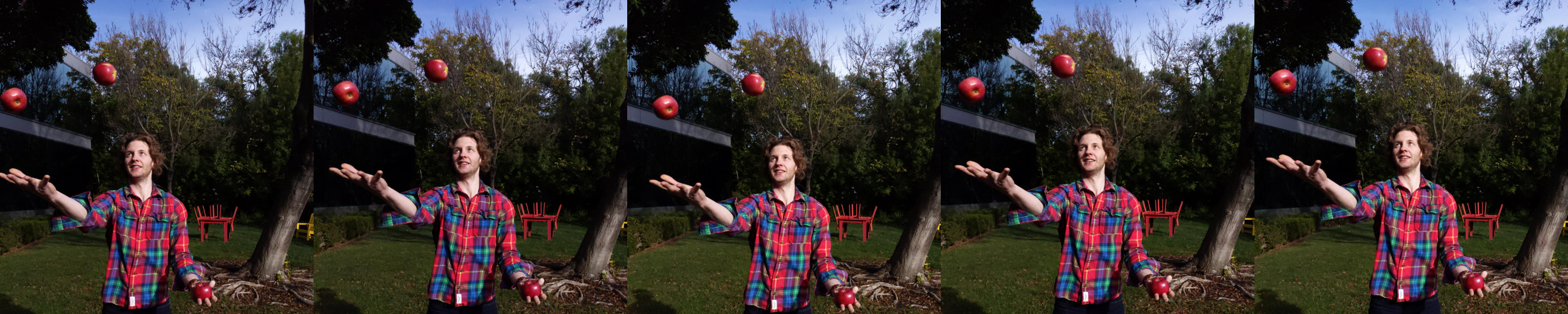} \\ \vspace{0.2mm}
    \caption{Additional captures from five synchronized smartphones. Note the consistent motion blur across images.}
    \label{fig:collages}
\end{figure*}

While our system has many applications, we focus on two here -- stereo reconstruction and image compositing of dynamic scenes. In addition, we use our method to capture multiple views of highly dynamic scenes (Fig.~\ref{fig:collages} and the supplement).
Our system has also been used for training a machine learning algorithm shipping with a commercial smartphone~\cite{LearningDepthBlog18}.

\textbf{Stereo reconstruction of dynamic scenes} - Stereo techniques reconstruct 3D geometry from two or more photos of the same scene from different viewpoints \cite{Furukawa15}. Assuming known camera poses, these techniques typically first identify a scene point's projections in multiple photos and then estimates its depth by triangulation. A fundamental assumption is that the scene point is stationary across photos. Therefore, we need synchronized capture to apply these techniques to dynamic scenes.  

\begin{figure}
    \centering
    \includegraphics[width=\linewidth]{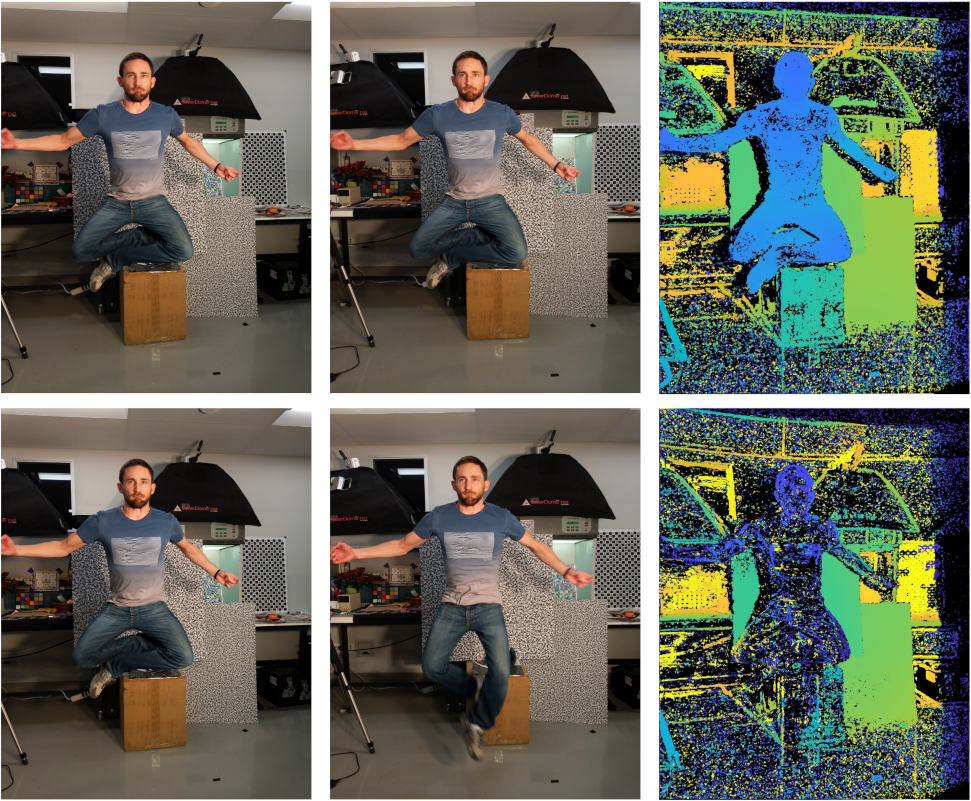}
    \caption{Top row: A stereo pair of a jumping subject captured using two smartphone cameras synchronized using our system and the corresponding depth map computed using COLMAP~\cite{COLMAP}. Bottom row: A stereo pair that is $\SI{66}{\milli\second}$ apart and the corresponding depth, representative of images one is likely to capture with a naive approach for synchronization.}
    \label{fig:depth}
\end{figure}

To qualitatively demonstrate the efficacy of our synchronization for depth estimation of dynamic scenes, we capture a pair of stereo images and compute depth using COLMAP \cite{COLMAP}, a state-of-art Structure-from-Motion \cite{SchonbergerSfm16} and Multi-View Stereo system \cite{SchonbergerMvs16} (Fig. \ref{fig:depth}). We compare it to naive synchronization, which we simulate by selecting frames that are $\SI{66}{\milli\second}$ apart, an amount that is comparable to the smallest average error among the different naive synchronization methods listed in Table~\ref{table:naive_vs_ours}. Clearly, improved synchronization results in higher quality depth. Moreover, since our approach scales to multiple cameras, we can further increase the accuracy of depth, especially around the occluded regions, by adding more synchronized cameras (see the supplement).

\textbf{Image compositing for dynamic scenes} - Image composites stitch multiple images of the same scene from different viewpoints into a single image\cite{Szeliski94, Szeliski97} that may otherwise be difficult or impossible to capture using a single camera, \eg, an extremely wide angle image or a multi-perspective image\cite{Agarwala06,Garg12}. A typical approach consists of aligning the images, projecting them onto a suitable compositing surface, and blending them to minimize alignment errors and account for differences in exposure, color balance, etc. 
While a composite may be constructed from images captured by a single camera, a dynamic scene requires synchronized captures from multiple cameras to minimize stitching artifacts.

\begin{figure}
    \centering
    \includegraphics[width=\linewidth]{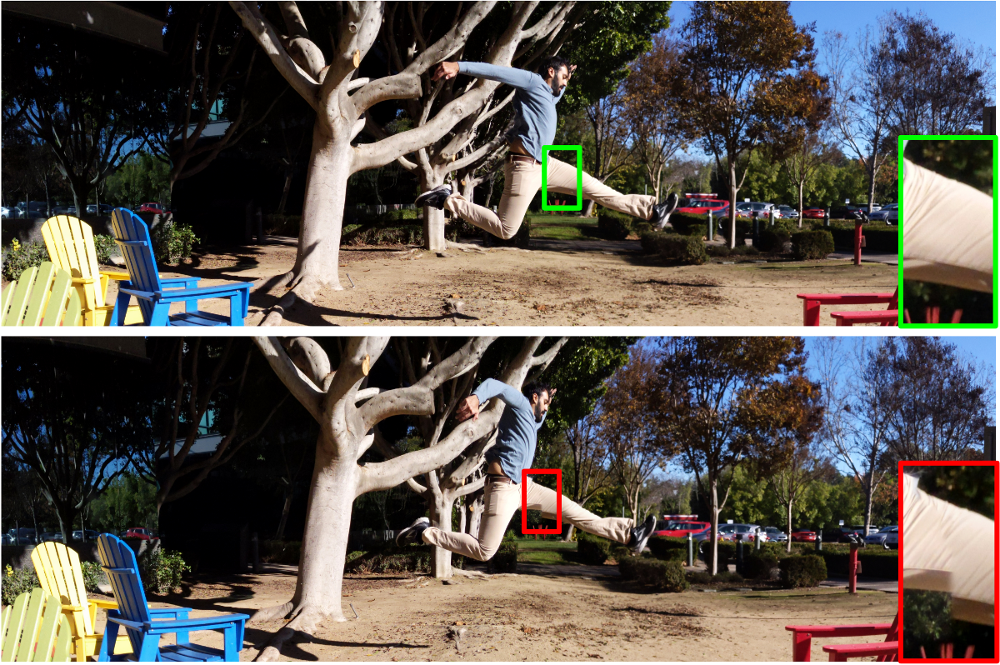}
    \caption{Composite of a dynamic scene stitched from two images using Microsoft Research's Image Composite Editor \cite{MSR_ICE}. When the two cameras are synchronized using our system, the resulting composite does not have artifacts (Top). However, a synchronization error of $\SI{66}{\milli\second}$ results in stitching artifacts (Bottom).}
    \label{fig:pano}
\end{figure}

Fig. \ref{fig:pano} demonstrates that our system can be used to avoid stitching artifacts due to motion. Our system can also synchronize exposure and white-balance across devices making it easier to blend between images in the composite.

\section{Conclusion and Future Work}
\label{sec:conclusion}
We presented an entirely software-based system for capturing time-synchronized image sequences on a collection of smartphones. Our method is wireless, runs on commodity hardware, and is able to achieve an accuracy of better than $\SI{250}{\micro\second}$, which we verified using a precision chronometer to be significantly better than naive methods that synchronize cameras prior to capture.


We empirically confirmed that the accuracy of our method largely depends on NTP clock bias, which can be minimized following the existing best practice of min filtering. We also analyzed and verified the convergence behavior of two phase alignment approaches: reset sampling, which is slow but widely applicable, and frame injection, which converges rapidly but has a minor hardware requirement.




As future work, we would like to scale up our system, \ie, to thousands of different devices distributed over a large geographical area maintaining synchronization over extended periods.
Our system has only been tested on Google Pixel 1, 2, and 3 smartphones.
The network is also restricted to $11$ devices (a limit imposed by the Android OS) and all clients must currently communicate directly with a designated leader device.
A peer-to-peer model can address both these limitations and adapt to both clock drift and motion. 
We open-source \texttt{libsoftwaresync} in hopes of inspiring new possibilities in social, collective photography applications.
We envision such a system being used to capture large-scale dynamic events such as concerts and sports matches.


\ifpeerreview
\else
\section*{Acknowledgment}
We thank Sam Hasinoff, Jon Barron, Roman Lewkow and Ryan Geiss for their helpful comments and suggestions, as well as Nikhil Karnard, Tim Brooks, Orly Liba, and David Jacobs for their help with collecting qualitative results.
\fi


\bibliographystyle{IEEEtran}
\bibliography{iccp19_bib}

\end{document}


\ifpeerreview
  \linenumbers
  \linenumbersep 5pt\relax
\fi 

%
\title{Supplement to Wireless Software Synchronization of Multiple Distributed Cameras}


\ifpeerreview
\author{Anonymous ICCP 2019 submission \\
Paper ID \paperID}
\else
\author{\IEEEauthorblockN{Sameer Ansari,
Neal Wadhwa, 
Rahul Garg, 
and Jiawen Chen}
\IEEEauthorblockA{Google Research, Mountain View, CA}
}
\fi

\ifpeerreview
\markboth{Anonymous ICCP 2019 submission ID \paperID}%
{}
\else
\fi

\maketitle

\section{Analysis of Clock Filters}

We present additional analysis of the mean and min clock filters.
We collected a dataset of 10,000 NTP messages together with a pair of images of our LED panel.
Since each camera is hardware timestamped in its device's local clock domain and the LED panel serves as an external clock, our setup lets us compute a reference timestamp for each message and therefore measure one-way latency.

Recall that the NTP time offset estimate is defined as:
\begin{equation}
\theta = \frac{(t_1 - t_0) + (t_2 - t_3)}{2}.
\end{equation}

\begin{figure}[h]
    \centering
    \includegraphics[width=\linewidth]{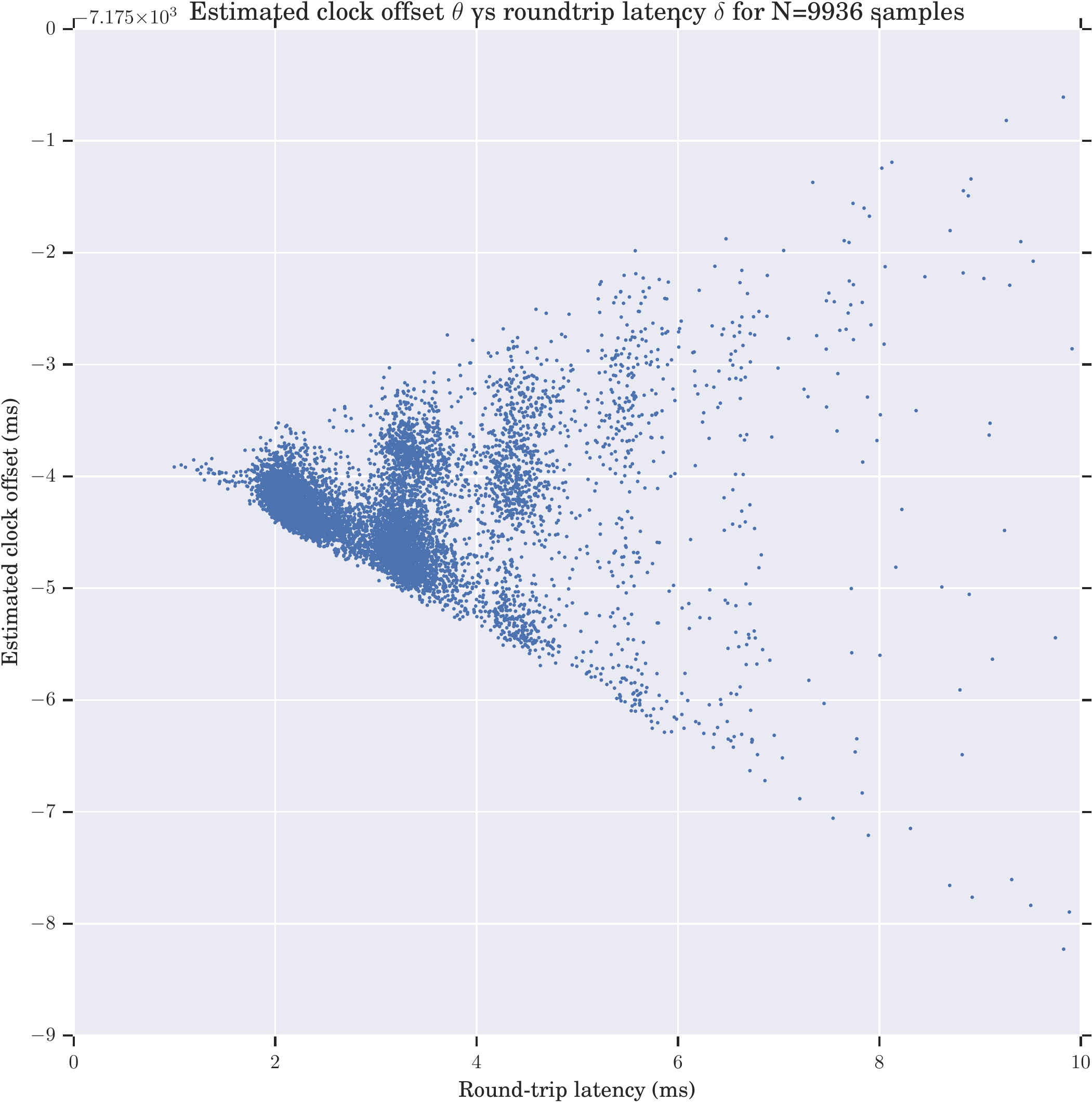}
    \caption{\emph{Wedge scattergram} of estimated clock offset $\theta$ on the $y$-axis and round-trip latency $\phi$ on the $x$-axis. Since higher latency samples have more variance when estimating $\theta$, the min filter looks for the apex of the wedge with lowest latency and therefore clock offset variance.}
    \label{fig:wedge_scattergram}
\end{figure}

Fig.~\ref{fig:wedge_scattergram} is a \emph{wedge scattergram}~\cite{Mills:2010:CNT:1951866}, plotting estimated clock offset $\theta$ vs. estimated round-trip latency $\phi$.
Notice that samples with larger round-trip latency also have greater variance in clock offset.
Since we can measure one-way latency, we can confirm the NTP hypothesis that the ``limbs'' of the wedge correspond to exchanges where the send and receive messages experienced the greatest difference in latency.
This observation suggest the use of a filter that selects for the samples with the lowest latency.
In our dataset of 10,000 messages, we observed that client to leader latency was consistently higher than the reverse by $\Delta \approx \SI{0.75}{\milli\second}$. If we compute $\theta$ for each sample in our dataset and robustly compute the mean by rejecting outliers with round-trip latency greater than $\SI{10}{\milli\second}$, we converge to a bias of $\Delta/2 \approx \SI{0.4}{\milli\second}$.

The min filter performs well if we ``get lucky'' and observe a message with low round-trip latency. Over 10,000 messages, the one with the lowest round-trip latency took $\SI{479}{\micro\second}$ from client to leader and $\SI{517}{\micro\second}$ from leader to client. Such a short period of time does not permit much buffering to contaminate our clock estimate, leading to a bias of only $\SI{19}{\micro\second}$. In our simple implementation of min filtering, once we find a good sample, the value persists until a sample with lower latency is later observed. It is straightforward to generalize to a more sophisticated strategy that uses the best $k$ samples, have old messages expire, etc.

\begin{figure}[!hb]
    \centering
    \includegraphics[width=\linewidth]{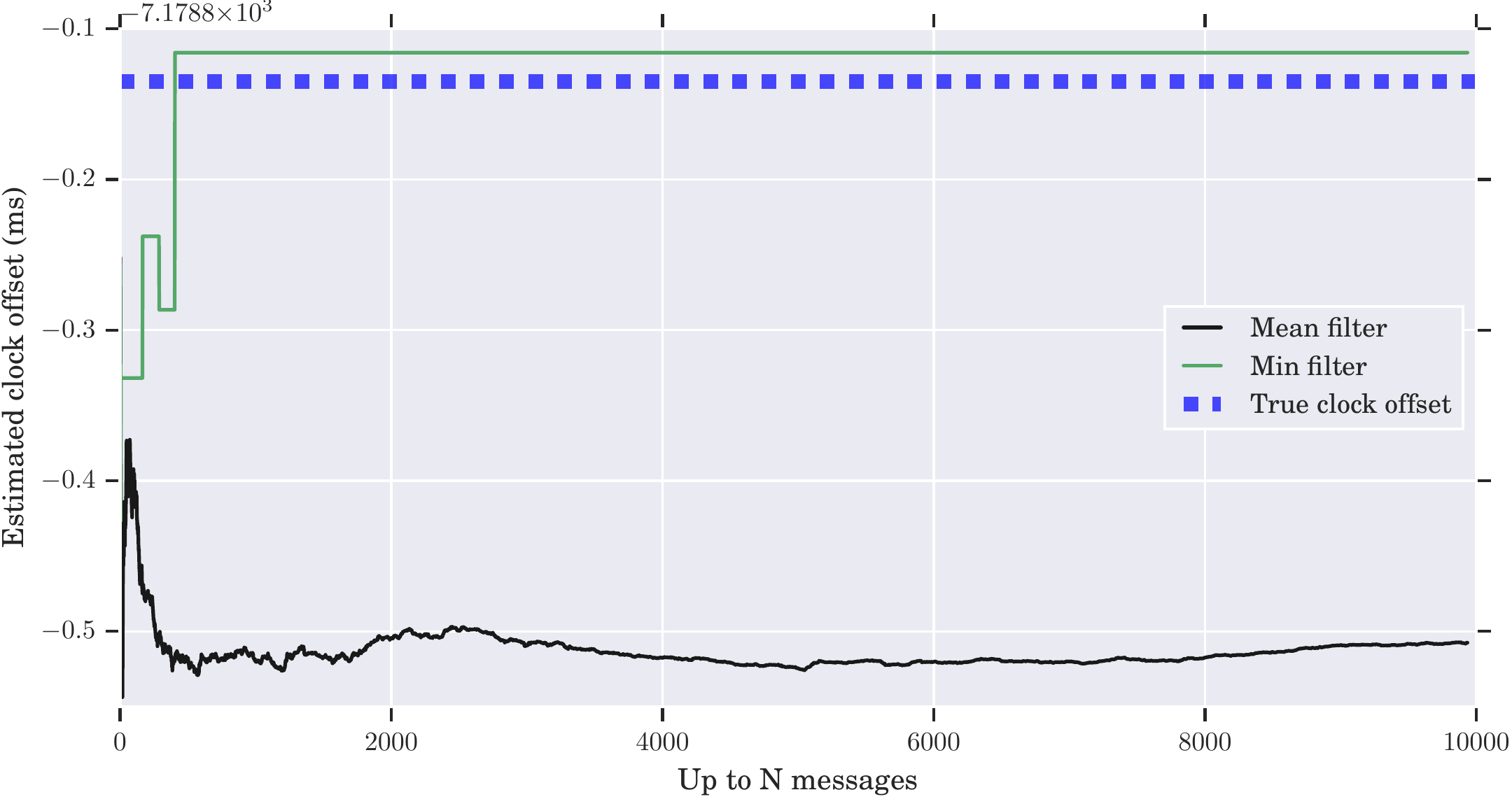}
    \caption{Convergence of mean vs. min filters. The mean filter converges to a systematic bias of approximately half the difference in one-way network latency. In our setup, the min filter finds a low-latency sample within $K=300$ messages.}
    \label{fig:min_vs_mean_convergence}
\end{figure}

The behavior of both mean and min filters accumulated over increasing sample counts is shown in Fig.~\ref{fig:min_vs_mean_convergence}. The mean filter converges to the bias of $\Delta/2$ while the min filter effectively latches onto the best sample found so far.




\section{Additional Results}

\subsection{Multi-view Stereo Results}

\begin{figure*}[!ht]
    \centering
    \includegraphics[width=\linewidth]{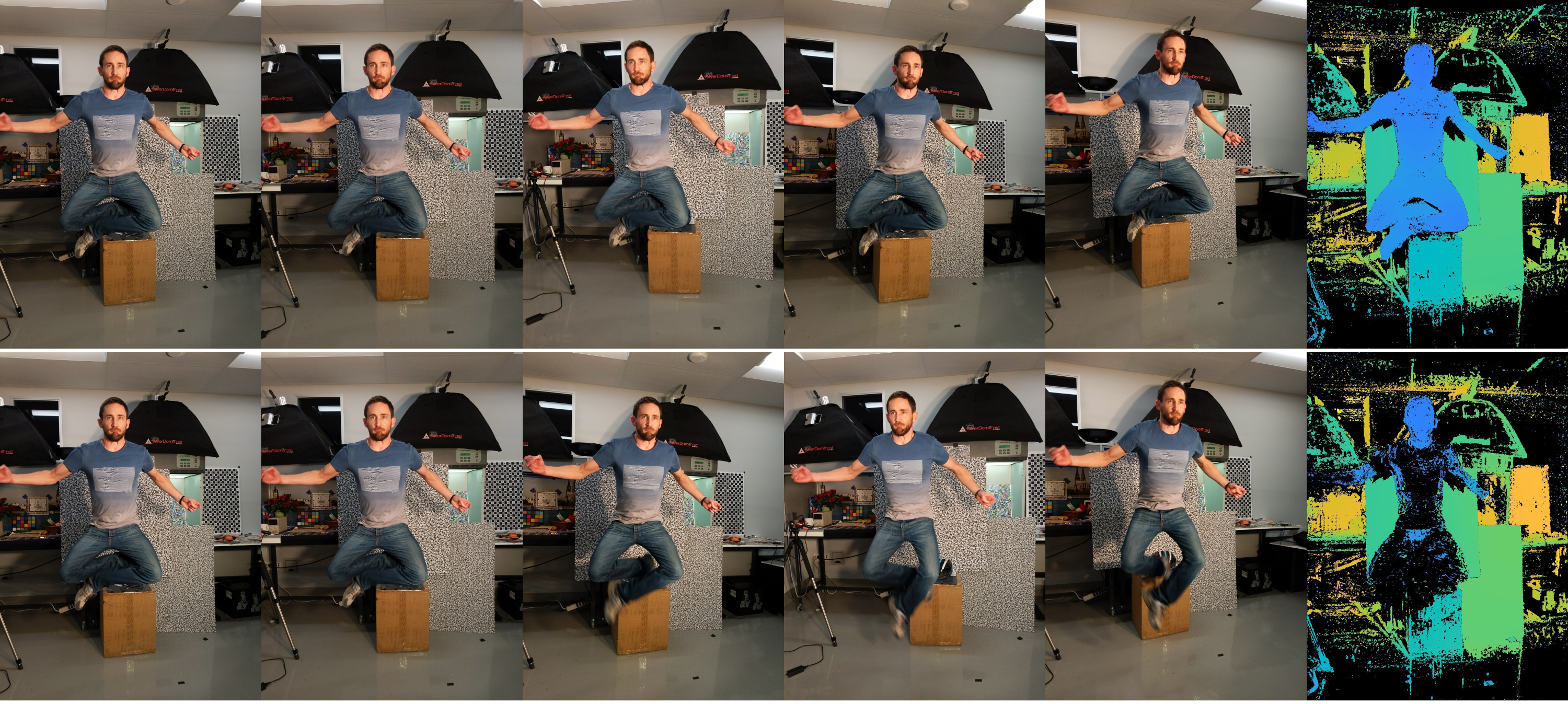}
    \caption{Multi-view stereo result for a jumping subject. Top: When the cameras are synchronized, the algorithm is able to use the information from all five views to generate a depth map that is better than the depth map we obtain from a stereo pair (Fig. 8 in the main paper). Bottom: When the views are not perfectly synchronized, e.g., when using a naive synchronization method, the depth map has missing information.}
    \label{fig:depth_supp_1}
\end{figure*}

\begin{figure*}[!ht]
    \centering
    \includegraphics[width=\linewidth]{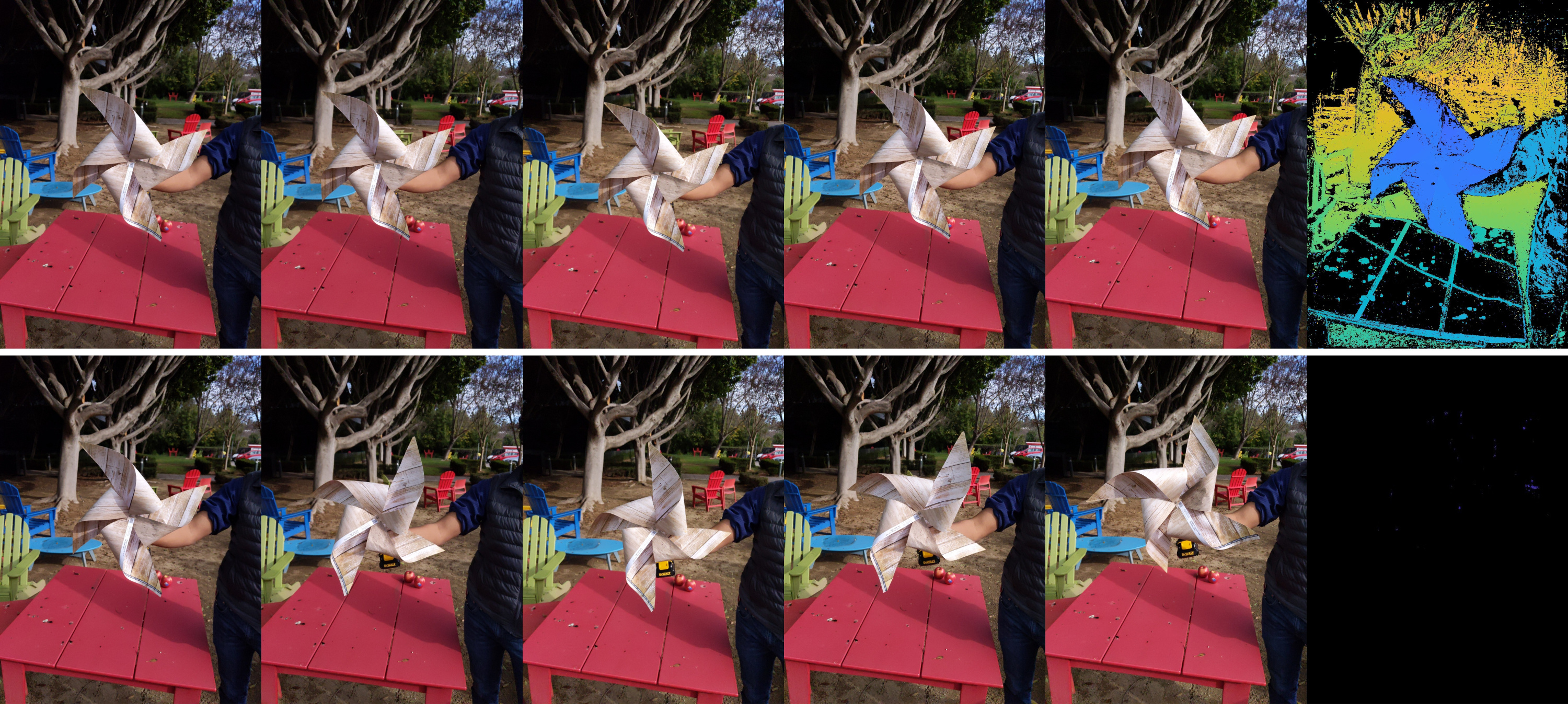}
    \caption{
    Multi-view stereo result for a rotating pinwheel when the cameras are synchronized using our system (Top) vs simulated naive synchronization (Bottom). Pose estimation fails in the latter case due to large differences in images.}
    \label{fig:depth_supp_2}
\end{figure*}

Multi-view stereo algorithms yield better depth maps as we add input views.
Since our system scales to multiple cameras, we can capture multiple synchronized views of dynamic scenes. Figures \ref{fig:depth_supp_1} and \ref{fig:depth_supp_2} show two such examples, comparing results when images are synchronized using our algorithm against the naive baseline. To simulate naive synchronization, we choose an offset for each camera sampled uniformly from  $\{\SI{-66}{\milli\second}, \SI{-33}{\milli\second}, \SI{0}{\milli\second}, \SI{+33}{\milli\second}, \SI{+66}{\milli\second}\}$ as measured from the leader. Depth maps are computed using COLMAP\cite{COLMAP}.


\subsection{Synchronized Multi-view Bursts}

\begin{figure*}[!ht]
    \centering
    \includegraphics[width=\linewidth]{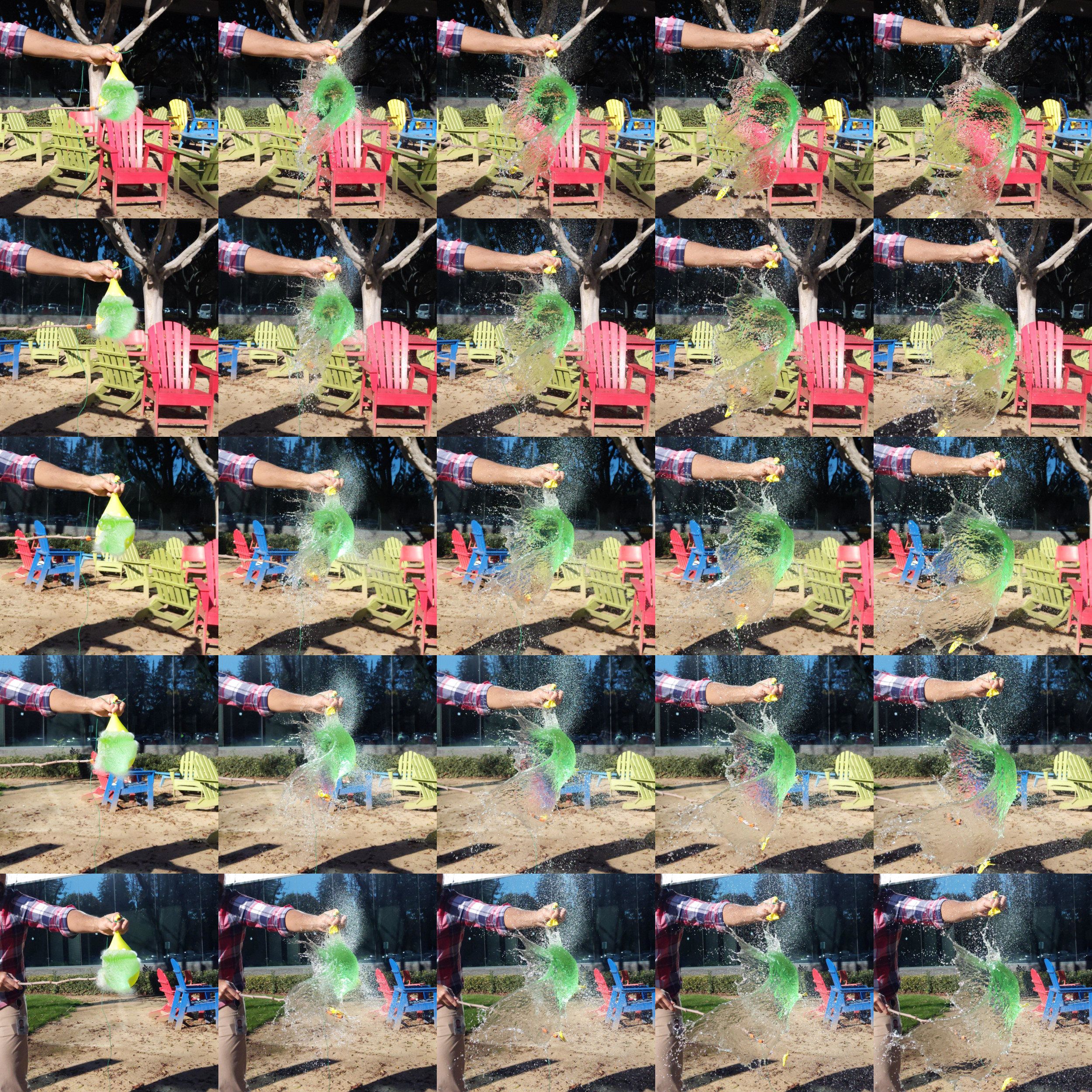}
    \caption{Synchronized bursts of a popping water balloon from five cameras. Each row is captured by the same camera while each column is a synchronized instant in time. Since there is sufficient light outdoors, we can use a short exposure time of $\SI{43}{\micro\second}$ while sensor readout time is $\SI{33}{\milli\second}$.}
    \label{fig:green_loaf}
\end{figure*}

Our system can capture high-resolution image sequences (``bursts'') or lower-resolution videos from multiple cameras and is limited only by disk bandwidth. Fig.~\ref{fig:green_loaf} shows a space-time volume of 5 frames from 5 views of a dynamic scene captured using our system.

\bibliographystyle{IEEEtran}
\bibliography{iccp19_bib}